\pdfoutput=1

\documentclass[11pt]{article}

\usepackage[final]{emnlp2021}

\usepackage{times}
\usepackage{latexsym}

\usepackage[T1]{fontenc}

\usepackage[utf8]{inputenc}

\usepackage{microtype}

\usepackage{algorithm}  
\usepackage{algorithmicx}  
\usepackage{algpseudocode}
\usepackage[utf8]{inputenc}
\usepackage{subfigure}
\usepackage{booktabs}
\usepackage{amsmath}
\usepackage{amsfonts}
\usepackage{multirow}
\usepackage{color}
\usepackage{enumitem}
\usepackage{microtype}
\usepackage{tabularx}

\usepackage{chngpage}
\usepackage{array}
\usepackage{microtype}
\usepackage{xcolor}
\usepackage{fontenc}
\usepackage{algpseudocode}
\usepackage{booktabs}
\usepackage{graphicx} 
\title{LUNA: Learning Slot-Turn Alignment for Dialogue State Tracking}

\author{Yifan Wang$^*$, Jing Zhao\thanks{~~Equal contribution.}, Junwei Bao\thanks{~~Corresponding author: baojunwei001@gmail.com}, Chaoqun Duan, Youzheng Wu, Xiaodong He \\
JD AI Research, Beijing, China \\
\{wangyifan15,zhaojing857,baojunwei,duanchaoqun1,wuyouzheng1,xiaodong.he\}@jd.com \;
}
    
\begin{document}
\maketitle
\begin{abstract}
Dialogue state tracking (DST) aims to predict the current dialogue state given the dialogue history. 
Existing methods generally exploit the utterances of all dialogue turns to assign value for each slot.
This could lead to suboptimal results due to the information introduced from irrelevant utterances in the dialogue history, which may be useless and can even cause confusion.
To address this problem, we propose \textbf{LUNA}, a S\textbf{L}ot-T\textbf{U}r\textbf{N} \textbf{A}lignment enhanced approach. It first explicitly aligns each slot with its most relevant utterance, then further predicts the corresponding value based on this aligned utterance instead of all dialogue utterances. 
Furthermore, we design a slot ranking auxiliary task to learn the temporal correlation among slots which could facilitate the alignment. 
%
Comprehensive experiments are conducted on multi-domain task-oriented dialogue datasets, i.e., MultiWOZ 2.0, MultiWOZ 2.1, and MultiWOZ 2.2. The results show that LUNA achieves new state-of-the-art results on these datasets.\footnote{Our code is available at \url{https://github.com/nlper27149/LUNA-dst}}


\end{abstract}



\section{Introduction}
Dialogue State Tracking (DST) refers to the task of estimating the dialogue state (\textit{i.e.,} user's intents) at every dialogue turn, where the state is represented in forms of a set of slot-value pairs \cite{williams2016dialog,eric2019multiwoz}.
DST is crucial to the success of a task-oriented dialogue system as the dialogue policy relies on the estimated dialogue state to choose actions.
Traditional DST approaches assume that all candidate slot-value pairs are predefined in an ontology~\cite{mrkvsic2016neural,zhong2018global,lee2019sumbt}.
Then, they scores all possible pairs 
and selecting the value with the highest score as
the predicted value of a slot. 


\begin{table}[h]
    \centering
     \small
    \begin{tabularx}{ \linewidth }{p{0.6cm} X}
        \toprule
         \textbf{Sys}: & There are lots to choose from. What type of cuisine are you looking for?  \\
        \textbf{User}:& I do not care. It needs to be on the \textcolor{green}{\textbf{south}} side and moderately priced. \\
        \textbf{State}: & \emph{restaurant-area=south}; \emph{pricerange=moderate} \\
        \midrule
        \textbf{Sys}: & There are 2 options, pizza hut cherry hinton and restaurant alimentum. Can I book you for those ? \\
        \textbf{User}: & Yes please. I also need a hotel with at least 3 stars and free parking near by the restaurant.\\
        \textbf{State}: & \emph{hotel-parking=yes}; \emph{hotel-stars
=3}\\
        \midrule
        \textbf{Sys}: & I am sorry, there is no guest house that meets those criteria, either. Would you like to try a different rating, or a different area?\\
        \textbf{User}: & Sure, what about in the city \textcolor{green}{\textbf{centre?}} \\
        \textbf{State}: & \emph{\textbf{hotel-area
=centre}}; \emph{hotel-type=guest house} \\
        ~~~~~~\textbf{$\bigotimes$}: & \emph{\textcolor{red}{\textbf{hotel-area=south}}}; \emph{hotel-type=guest house}\\
        \bottomrule
    \end{tabularx}
    \caption{\label{example} An example of DST. ``\textbf{User}" and ``\textbf{Sys}" means user query and system response respectively. ``\textbf{State}" is the golden label of dialogue state. ``$\bigotimes$" denotes the predicted states of some existing models and the state marked \textcolor{red}{red} is the incorrect prediction.}
\end{table}

DST encounters many challenging phenomena unique to dialogue, such as co-references and ellipsis. 
Consequently, most of existing DST approaches exploit all dialogue utterances in history to assign value for each slot~\cite{shan2020contextual,chen2020parallel,quan2020modeling,hu2020sas,chen2020schema}.
However, this could lead to the incorrect value assignment due to the ambiguous contents  introduced from some irrelevant utterances with the current slot.
As the example shown in Table~\ref{example}, the models estimate a slot value ``\textit{south}"  for the slot ``\textit{hotel-area}" at turn-3 yet its corresponding golden label is ``centre". 
The reason is that both \textit{``south"} and \textit{``centre"} are the potential slot values to the area-related slot (\textit{i.e.,} ``\textit{restaurant-area}" and ``\textit{hotel-area}") in the ontology. 
Actually,  the domain of the utterance at turn-1  is ``\textit{restaurant}" that is irrelevant to the slot ``\textit{hotel-area}".


To address the problem aforementioned, we propose \textbf{LUNA}, a S\textbf{L}ot-T\textbf{U}r\textbf{N} \textbf{A}lignment enhanced approach,  which divides DST into two sub-tasks: (1) explicitly aligns each slot with its most relevant utterance, (2) assigns the slot value according to the aligned utterance.
For example, when predicting the slot value of ``\textit{hotel-area}", LUNA first aligns it with the relevant utterance (\textit{i.e.,} turn-3) and then only uses the representations of this utterance to match slot value.
Concretely, LUNA consists of four parts: an utterance encoder, a slot encoder, a value encoder, and an alignment module between the first two encoders.
The core of LUNA is the alignment module directed at accurate alignment, otherwise there may be a risk of the failure of the second sub-task.
Correspondingly, the alignment module equipped in LUNA is performed by an iteratively bi-directional feature fusion network based on the attention mechanism.
Some previous works have explored the feature fusion of the two encoders, but they are all uni-directional~\cite{shan2020contextual,chen2020schema,ye2021slot}, \textit{e.g.,} turn-to-slot or slot-to-turn. Compared with them, the bi-directional 
way can build a mutual relevance between two encoders and thus more effective for our alignment-oriented objective.


Additionally, we design a ranking-based auxiliary task to supervise LUNA to learn the slot order along with the conversational flow, which could facilitate the alignment. 
For example, the order of the slots in Table~\ref{example} is:

\begin{small}
\noindent   (1) ``\textit{restaurant-area}" (2) ``\textit{pricerange}" (3) ``\textit{hotel-parking}"

\noindent (4) ``\textit{hotel-stars}" ~~~~~~~~(5) {\color{red}``\textit{hotel-area}"} ~(6) ``\textit{hotel-type}"
\end{small}

\noindent Among the above slots, the most difficult-aligned slot is ``\textit{hotel-area}" which confronts the confusion from the utterances at turn-1 (containing ``\textit{south}") and turn-3 (containing ``\textit{center}").
But the remaining five slots are easy-aligned, such as  ``\textit{hotel-stars}".
If the model combines two information: (1) ``\textit{hotel-stars}" is aligned with the  utterance at turn-2, (2) the conversation order of ``\textit{hotel-area}" is after ``\textit{hotel-stars}",
it can easily inference that ``\textit{hotel-area}" should be aligned with the  utterance at turn-3.
Notably, our proposed auxiliary task enables LUNA to learn the semantic correlations as well as the temporal correlations among slots. 
Whereas, existing DST approaches only attempt to model the semantic correlations~\cite{ye2021slot,zhu2020efficient,chen2020schema}.


Comprehensive experiments are conducted and the results show that LUNA achieves state-of-the-art (SOTA) on three of the most actively studied datasets: MultiWOZ 2.0 \cite{budzianowski2018multiwoz}, MultiWOZ 2.1 \cite{eric2019multiwoz}, and MultiWOZ 2.2 \cite{zang2020multiwoz} with joint accuracy of 55.31\%, 57.62\%, and 56.13\%. The results outperform the previous SOTA by +0.97\%, +1.26\%, and +4.43\%, respectively. Furthermore, a series of subsequent ablation studies demonstrate the effectiveness of 
each module in our model. 
Our main contributions are summarized as follows:

(1) We propose a DST approach LUNA which mitigates the problem of incorrect value assignment through explicitly aligning each slot with its most relevant utterance.

(2) We propose an auxiliary task to facilitate the alignment which is firstly introduced in DST to take the temporal correlations among slots into account.

(3) Empirical experiments are conducted to show that LUNA achieves SOTA results with significant improvements.

\section{Related Work}
DST is a necessary component in task-oriented dialogue systems and a large amount of work has been proposed to achieve better performance. All these methods can be broadly divided into two categories: classification \cite{xu2018end,zhong2018global,ren2018towards,xie2018cost} and generation \cite{wu2019transferable,hosseini2020simple,kim2019efficient}.
The classification method requires that all possible slot-value pairs are given in a predefined ontology.
Then, the pair with the highest score is the final prediction. 
Conversely, the generation way does not rely on manual definition, which generates dialogue states from utterances using the seq2seq fashion. 
This work is mainly related to the classification method.

Recently, transformer-based pre-trained models, such as BERT \cite{devlin2018bert}, have achieved remarkable results in a range of natural language processing tasks.
Thereupon, the research of DST has been shifted to building new models on top of the powerful pre-trained language models.
SUMBT~\cite{lee2019sumbt} is the first model to employ BERT to model the relationships between slots and dialogue utterances through a slot-word attention mechanism. 
CHAN \cite{shan2020contextual} presents a hierarchical attention network which uses slot-word attention and slot-turn attention to enhance the representations of slots.
All the methods mentioned above predict the value of each slot separately and ignore the correlations among slots.
SST \cite{chen2020schema} incorporates graph attention networks into DST and proposes schema graphs which contain slot relations in edges.
STAR \cite{ye2021slot} provides a slot self-attention mechanism to learn mutual guidance among slots and enhance the ability to deduce appropriate slot values from related slots.
Recently, BORT~\cite{BORT} proposes a reconstruction mechanism which enhances the performance of DST.

To the best of our knowledge, we are the first to reveal that exploiting all dialogue utterances to assign value may cause suboptimal results and the first to learn the temporal correlations among slots.

\section{Methodologies}
\begin{figure*}[tb]
	\centering
	\includegraphics[width=14cm]{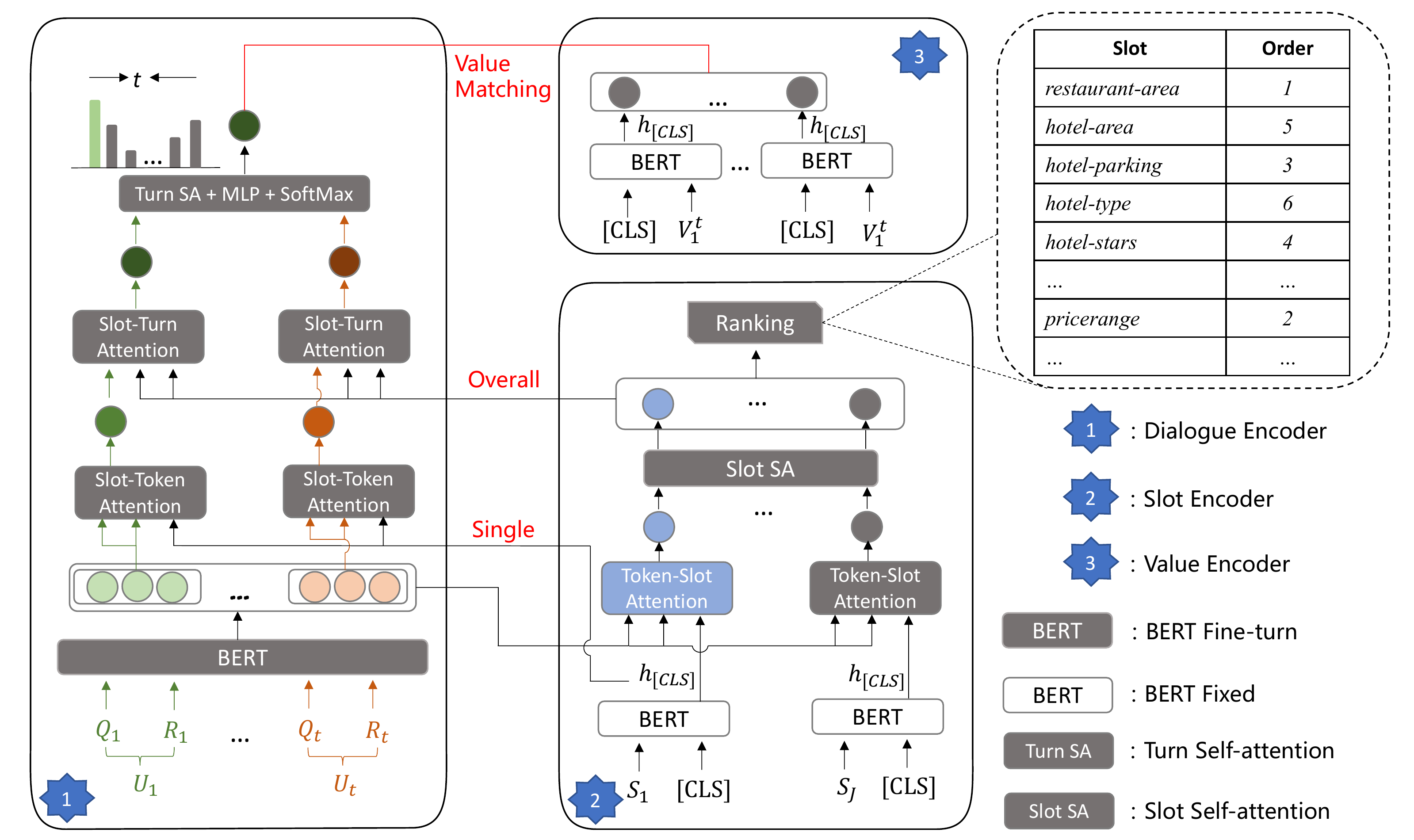}
	\caption{\label{fig:TSA-DST} The architecture of LUNA. Note that the workflow in this figure is specifically for the first slot $S_1$. }
	\vspace{-0.3cm}
\end{figure*}

Suppose that there is a conversation composed of $T$ utterances, $\mathcal{X}=\{(Q_{1},R_{1}),...,(Q_{T},R_{T})\}$,
and a predefined slot set $\mathcal{S}=\{S_{1},...,S_{J}\}$, 
where $Q_{t}$ denotes the user query at $t$-th utterance, $R_{t}$ is the corresponding system response and $J$ is the total number of slots.
DST aims to predict states at each turn with given utterances up-to-now 
$(Q_{\leq t},R_{\leq t})$,
and presents them as slot-value pairs, $\mathcal{B}_{t}=\{(S_{1},V_{1}^{t}),...,(S_{J},V_{J}^{t})\}$,
where $S_{j}$ is the $j$-th slot in $\mathcal{S}$ ,
and $V_{j}^{t}$ is the value with respect to $S_{j}$ for the $t$-th turn.
Since the datasets are collected from multi domains, following previous works~\cite{hu2020sas,kim2019efficient},
we concatenate domain names and slot names as domain specific slots.

To tackle this task, we propose the LUNA model.
As depicted in Figure \ref{fig:TSA-DST}, this model consists of three encoders and an alignment network.
In this section, we will elaborate each module of this model.


\subsection{Encoders}
Inspired by the success of the pre-trained model in the community of the NLP, we adopt the BERT~\cite{devlin2018bert} to implement the context encoder.

\subsubsection{Utterance Encoder}
Given the $t$-th utterance $(Q_{t}, R_{t})$ and its history 
$(Q_{\leq t},R_{\leq t})$,
we first concatenate them into a single sequence: $\mathcal{X}_{t}=Q_{1}\oplus R_{1}\oplus \cdots \oplus Q_{t}\oplus R_{t}$.
Following the form of the input of the BERT, we then surround the sequence with two special tokens [CLS] and [SEP].
Given that not all of the slots can be aligned to a specific utterance, we further add an extra token [BLANK] as a placeholder. 
All of the slots that are not mentioned in the dialogue are aligned to [BLANK].
Finally, the input of the utterance encoder can be denoted as follows:
\begin{align}
X_t = \text{[CLS]} \oplus \mathcal{X}_{t} \oplus \text{[SEP]} \oplus \text{[BLANK]}.
\end{align}

After obtaining $X_{t}$, we feed it into the BERT to learn semantic representations:
\begin{align}
\mathbf{H}_{t}=\text{BERT}_{finetune}(X_{t}),
\end{align}
where $\mathbf{H}_{t}=[\mathbf{h}_{1}^{t},\mathbf{h}_{2}^{t},...,\mathbf{h}_{|X_{t}|}^{t}]$, $\mathbf{h}_{j}^{t}\in \mathbb{R}^{d}$.
Additionally, we add a learned embedding to every token indicating which turn it belongs to. Thus, for a given token in $X_{t}$, its input representation is constructed by summing the corresponding token, position, segment, and turn embeddings. In order to make the BERT more adapt to this task, we fine tune the parameters of the BERT during the training stage.

\subsubsection{Slot and Value Encoders}
Following previous works~\cite{shan2020contextual,ye2021slot}, we leverage another BERT to encode slots and their candidate values. 
Formally, given a slot $S_{j}$ or a value $V_{j}^{t}$, we first tokenize it into a sequence and then concatenate it with the special token [CLS] to build the input for the slot or value encoder.
After that, we exploit the BERT to encode the concatenation as follows:
\begin{align}
\mathbf{h}_{s_j} &= \text{BERT}_{fixed}(S_j), \\
\mathbf{h}_{v_{j}^{t}} &= \text{BERT}_{fixed}(V_{j}^{t}).
\end{align}
We regard the representation of [CLS] as that of the whole slot or value.
Specially, since the quantity of the sub-vocabulary related to slots and values are small,
we freeze the parameters of BERT in slot and value encoders during the training stage.

\subsection{Alignment Module}
As mentioned above, DST model usually adopts all of previous utterances as the history to enhance the representation of the current utterance.
Although this mechanism enriches the semantic representation, it introduces some noisy and causes confusion for value prediction to a specific slot.
To alleviate this issue, the proposed LUNA model adopts iteratively bi-directional feature fusion layers, turn-to-slot and slot-to-turn, to align slots to utterances and provide more relevant utterance for value prediction.
\subsubsection{Turn-to-Slot Alignment}
In this work, we regard utterances and slots as two sequences and aims to align them with each other.
To this end, we first employ a multi-head attention mechanism~\cite{vaswani2017attention} to assist the slots extracting relevant information from utterances based on the outputs of the utterance and slot encoders:
\begin{align}
\mathbf{h}_{s_{j},t} = \text{MultiHead}(\mathbf{h}_{s_j}, \mathbf{H}_{t}, \mathbf{H}_{t}),
\label{token-slot attention}
\end{align}
where $\text{MultiHead}(\cdot,\cdot,\cdot)$ denotes the multi-head attention mechanism.
Through this operation, we obtain utterance-aware slot representations.

After that, we adopt $N$ stacked layers to learn the correlation among slots,
and each layer consists of a multi-head self-attention mechanism and a position-wise feed-forward network.
We denote this module as Slot SA.
Formally, the $n$-th layer is computed as follows:
\begin{align}
\bar{\textbf{H}}_{s}^{n}&=\text{MultiHead}(\hat{\textbf{H}}_{s}^{n-1},\hat{\textbf{H}}_{s}^{n-1},\hat{\textbf{H}}_{s}^{n-1}), \\
&\hat{\textbf{H}}_{s}^{n}=\text{FNN}(\text{ReLU}(\text{FNN}(\bar{\textbf{H}_{s}^{n}}))
\label{slot}
\end{align}
where $\hat{\textbf{H}}_{s}^{1}=[\textbf{h}_{s_{1},t},...,\textbf{h}_{s_{J},t}]$.

\subsubsection{Slot-to-Turn Alignment}
For utterances, after obtaining the output of the utterance encoder $\textbf{H}_{t}$,
we first slice it into $t$ segments, $\textbf{U}=[\textbf{U}_{1},...,\textbf{U}_{t}]$ and each segment corresponds to an utterance.
We then exploit a hierarchical attention mechanism to model the slot-to-turn alignment.
The hierarchical attention mechanism contains two layers.
The first layer models the preliminary alignment between an utterance and a slot and we denote it as \textbf{Single Slot-to-Turn}.
The other one focuses on the refined alignment through incorporating all slots information and we represent it as \textbf{Overall Slot-to-Turn}.

As shown in Figure~\ref{fig:TSA-DST}, the Single Slot-to-Turn is responsible for extracting token-level information related to a specific slot from each utterance.
Take the $j$-th slot $S_{j}$ as an example.
Given its representation $\mathbf{h}_{s_{j}}$,
we use it to extract most relevant information from each utterance (\textit{e.g.,} $i$-th utterance) via the multi-head attention mechanism:
\begin{align}
\bar{\textbf{U}}_{i}=\text{MultiHead}(\mathbf{h}_{s_{j}}, \textbf{U}_{i}, \textbf{U}_{i}),
\label{single}
\end{align}
where $\bar{\textbf{U}}_{i}$ is a $d$-dimension vector and we regard it as slot $S_{j}$ aware representation for $i$-th utterance.
Similarly, we obtain slot $S_{j}$ aware representations for all utterances $\bar{\textbf{U}}=[\bar{\textbf{U}}_{1},...,\bar{\textbf{U}}_{t}]$ with the same operation.

After that, the Overall Slot-to-Turn layer further aligns utterances with slots.
Different with existing work \cite{ye2021slot} of encoding the states of the previous turn $B_{t-1}$ as an information supplement, we first introduce previous alignment information into each utterance by adding alignment embedding:
\begin{align}
\hat{\textbf{U}}_{i}=\bar{\textbf{U}}_{i}+\text{AE}(i),
\end{align}
where AE is embedding matrix indicating whether the slot $S_{j}$ aligns utterance $\textbf{U}_{i}$ or not at last turn. Then we utilize another multi-head attention module to update utterance representations based on slots information as follows:
\begin{align}
\tilde{\textbf{U}}_{i}=\text{MultiHead}(\hat{\textbf{U}}_{i},\hat{H}_{s}^{N},\hat{H}_{s}^{N}).
\label{turn-slot}
\end{align}

To aggregate context dependency among utterances, we further introduce a multi-head self-attention mechanism to learn the context aware representation for each utterance:
\begin{align}
\textbf{D}=\text{MultiHead}(\tilde{\textbf{U}},\tilde{\textbf{U}},\tilde{\textbf{U}}),
\label{1111}
\end{align}
where $\tilde{\textbf{U}}=[\tilde{\textbf{U}}_{1},...,\tilde{\textbf{U}}_{t}]$, $\textbf{D}=[\textbf{D}_{1},...,\textbf{D}_{t}]$.
$\textbf{D}$ is adopt to predict the alignment distribution over turns for $S_j$ as follows:
\begin{align}
p(\cdot|S_{j})=\text{softmax}(\textbf{W}_{o}\textbf{D}+b_{o}),
\label{softmax}
\end{align}
where $\textbf{W}_{o}\in \mathbb{R}^{d}$ and $b_{o}$ are trainable parameters.
We employ the cross-entropy as the objective function of the alignment and it can be formulated as:
\begin{align}
\mathcal{L}_{align}=-\sum_{j=1}^{J}\log p((Q^*_j,R^*_j)|S_{j}),
\end{align}
where $(Q^*_j,R^*_j)$ is ground-truth utterance aligned to slot $S_{j}$.


\subsubsection{Auxiliary Ranking Task} The output of the Slot SA, $\hat{\textbf{H}}_{s}^{N}=[\hat{\textbf{h}}_{s_{1}}^{N},...,\hat{\textbf{h}}_{s_{J}}^{N}]$, only contains the information of the semantic correlations among slots.
To facilitate the alignment, the model needs the assistance of the temporal correlations among slots.
However, slots are naturally disordered or sorted in lexicographic order.
Therefore, we design an auxiliary task to guide the model to learn the temporal information of slots.
Particularly, we propose an ordering algorithm to determine the slots order with respect to the dialogue utterances, as shown in Algorithm~\ref{order}.
%
This task aims to minimize the order differences between the disordered slots and our defined-ordered slots and we utilize the ListMLE~\cite{xia2008listwise} as the objective function.
ListMLE is a standard ranking loss and it is computed based on a defined list and a ground-truth list.
To compute the loss, we learn a score for each slot (\textit{e.g.,} $S_{j}$) as follows:
\begin{align}
\textbf{f}_{s_{j}}=\text{Sigmod}(\mathbf{W}_s\hat{\textbf{h}}_{s_{j}}^{N}+b_s),
\end{align}
where $\mathbf{W}_s$ and $b_s$ are trainable parameters.
Given the ground-truth order of slots $\textbf{o}=[o_{1},...,o_{J}]$ and the corresponding slot list is $[S_{o_{1}},...,S_{o_{J}}]$, the loss function can be formulated as follows:
\begin{align}
p(j|S_{o_{j}})=\frac{\exp(f_{s_{o_j}})}{\sum_{l=j}^{J}\exp(f_{s_{o_l}})}, \\
\mathcal{L}_{order}=-\log(\prod_{j=1}^{J}p(j|S_{o_{j}})).
\end{align}

\begin{algorithm}[h]
  \caption{Slots Ordering Algorithm}
    {\bf Input: } 
  $ L$: Label slots for a conversation, $T$: the number of turns in this conversation \\
      {\bf Initialize: } 
  {\bf $S$} : A list of sorted slots 
  \begin{algorithmic}[1]
        \For{$t\in [1,T]$}
            \State Find the label slots $L_t=[l_{t,1},..., l_{t,n}]$ of $\mbox{$t$-th}$ turn;
            \State Sort $L_t$ by slots' lexicographic order;
            \For {$l$ in $L_t$}
                \State Add $l$ to $S$;
            \EndFor
        \EndFor
    \State Define the list of remaining not-aligned slots is $L_{blank}$;
    \State Sort $L_{blank}$ by slots' lexicographic order;
    \For {$l$ in $L_{blank}$}
        \State Add $l$ to $S$;
    \EndFor
  \end{algorithmic}
  \label{order}
\end{algorithm}

\subsection{Value Prediction}
Above sections describe the method of aligning slots with utterances.
We then predict the value for a specific slot based on the most relevant utterance
instead of all of the utterances.

Formally, given a slot $S_{j}$, we first select the most relevant utterance $(Q^*_j, R^*_j)$ as follows:
\begin{align}
(Q^*_j, R^*_j)=\arg\max (\{p((Q_{i},R_{i})|S_{j})\}_{i=1}^{t}).
\end{align}
Then we feed its representation $D^*\in D$ into a linear layer which is followed by a layer normalization:
\begin{align}
O^*=\text{LayerNorm}(\text{Linear}(D^*)).
\end{align}

Following \citet{ren2018towards}, we adopt the L2-norm to compute the distance between a slot and a candidate value.
Thereby, the value prediction probability distribution can be formulated as follows:
\begin{equation}
\begin{aligned}
    p(V^t_{j}|(Q_{\leq t},&R_{\leq t}),S_{j})  = \\
  & \frac{\exp(-\|O^{*}-h_{v_{j}^{t}}\|_{2})}{\sum_{V_{k}^{t}\in V^{t}} \exp(-\|O^{*}-h_{v_{k}}^{t}\|_{2})}, 
\end{aligned}
\end{equation}


where $V^{t}$ is the set of candidate value of slot $S_j$ for the $t$-th utterance.
Finally, the loss function can be defined as:
\begin{align}
\mathcal{L}_{value}=-\sum_{j=1}^{J}\log(p(V^t_{j}|(Q_{\leq t},R_{\leq t}),S_{j})).
\end{align}

\begin{table*}

\renewcommand{\arraystretch}{1.0} 
  \centering
  \fontsize{9}{9}\selectfont	

	\begin{tabular}{lclcclcc}
	    \toprule
		\multirow{2}*{Model} & \multicolumn{2}{c}{MultiWOZ 2.0} & \multicolumn{2}{c}{MultiWOZ 2.1} & \multicolumn{2}{c}{MultiWOZ 2.2} & \multirow{2}*{Trainable Parameters} \\
		\cmidrule(lr){2-3} \cmidrule(lr){4-5} \cmidrule(lr){6-7}
		& Joint & Slot & Joint & Slot & Joint & Slot\\
 		\midrule
 		\textbf{Generation models} &&&&&&& \\
 		SOM-DST~\cite{kim2019efficient} & 51.38 & - & 52.57 & - & - & - & 113M \\
 		TRADE~\cite{wu2019transferable} &48.60 & 96.92 & 45.60 & - & 45.40$^\dag$ & - & -\\
 		TripPy~\cite{heck2020trippy} & 53.51 & - & 55.32 & - & 53.52 & - & 110M\\
 		TripPy w/o LM & 45.64$^\dag$ & - & 44.80$^\dag$ & - & - & - & -\\ 		
 		Seq2Seq-DU~\cite{feng2020sequence} &  - & - & 56.10 & - & 54.40 & - & 220M \\
 		SimpleTOD~\cite{hosseini2020simple} & 51.37 & - & 51.89 & - & - & - & -\\
 		
 		\midrule  
 		\textbf{Classification models} &&&&&&&  \\
 		DS-DST~\cite{zhang2019find} & - & - & 51.31 & 97.35 & 51.70 & - & -\\ 		
 		DST-Picklist~\cite{zhang2019find} & 54.39 & - & 53.30 & 97.40 & - & - & -\\ 		
 		CHAN~\cite{shan2020contextual} & 53.06 & - & 53.38 & - & - & - & 133M \\
 		SST~\cite{chen2020schema}  & 51.17 & - & 55.23 & - & - &- & - \\
		STAR~\cite{ye2021slot} & 54.34 & - & 56.36 & 97.51 & - & - & 135M\\
        \midrule
		\textbf{LUNA} &\bf55.31 & \bf97.35 & \bf57.62 & \bf97.96 & \bf56.13 & \textbf{97.68} & 142M\\
		\midrule
 		\multicolumn{8}{c}{With Data Augmentation} \\
 		\midrule
 		TripPy+ConvBERT~\cite{mehri2020dialoglue} & -&-&58.70&-&-&- & - \\
 		TripPy+CoCoAug~\cite{li2020coco} & -&-&60.53&-&-&-& - \\
 		TripPy+SaCLog~\cite{dai2021preview} & -&-&60.61&-&-&- & -\\
	   \bottomrule
	\end{tabular}
	\caption{\label{main result}Joint accuracy (\%) and slot accuracy (\%) on the test sets. ``LM" denotes label map in TripPy. {\dag} indicates the reproduced results using the source codes and remaining results reported in the literature.  }
	\vspace{-2mm}
\end{table*}

\subsection{Optimization}
We adopt the multi-task learning to jointly optimize the alignment loss, value prediction loss and the auxiliary task loss. The total loss is defined as follows:
$$\mathcal{L}_{joint} = \mathcal{L}_{order}+\mathcal{L}_{align}+\mathcal{L}_{value}$$

\section{Experimental Setup}
\subsection{Datasets and Metrics}
We evaluate our approach on three gradually refined task-oriented dialogue datasets: MultiWOZ 2.0~\cite{budzianowski2018multiwoz}, MultiWOZ 2.1~\cite{eric2019multiwoz}, and the latest MultiWOZ 2.2~\cite{zang2020multiwoz}, containing over 10,000 dialogues, 7 domains, and 35 domain-slot pairs.
MultiWOZ 2.1 modifies about 32\% of the state annotations in MultiWOZ 2.0. 
MultiWOZ 2.2 is the latest and a further refined version of MultiWOZ 2.1, which solves the inconsistency of state updates and some problems of ontology.

We use joint accuracy and slot accuracy as our evaluation metrics. Joint accuracy is the proportion of dialogue turns where the value of each slot is correctly predicted. Slot accuracy only considers individual slot-level accuracy. The ground-truth of slot value is set to none if the slot has not been mentioned in dialogue. 

\subsection{Training}
Same as the previous work~\cite{shan2020contextual,ye2021slot}, we use BERT-base-uncased model as the encoders of LUNA where only the utterance encoder is fine-tuned and the parameters of the other two encoders are fixed. BERT-base has 12 layers of 784 hidden units and 12 self-attention heads.
The number of attention heads in multi-head attention in our alignment module is set to 4. 
The number of layers in slot self-attention and turn self-attention is set to 4 and 2 respectively. 
During the training process, we use Adam optimizer~\cite{kingma2014adam} and set the warmup proportion to 0.1. 
Considering that the encoder is a pre-trained BERT model
while the other parts in our model needs to be trained from scratch, we use different learning rates for those parts. 
Specifically, the peak learning rate is set to 3e-5 for the utterance encoder and 1e-4 for the remaining parts. 
The maximum input sequence length in BERT is set to 512. For MultiWOZ 2.0, MultiWOZ 2.1, and MultiWOZ 2.2, we apply the same hyperparameter settings.

\section{Experiment Results}
\subsection{Main Results}

Table \ref{main result} shows the joint accuracy and the slot accuracy of our model and other baselines on the test sets of MultiWOZ 2.0, 2.1, and 2.2, where some models are not tested on the 2.2 version since it was released shortly. As shown in the table, among the models without data augmentation, our model LUNA achieves state-of-the-art performance on these datasets with joint accuracy of 55.31\%, 57.62\%, and 56.13\%, which has a measurable improvement (0.97\%, 1.26\%, and 4.43\%) over the previous best results, illustrating the effectiveness of slot-turn alignment in DST task.

		

It can be observed that the three data augmented methods reach higher than 58\% joint accuracy on MultiWOZ 2.1.
We believe that these data augmentation skills are versatile.
If they can improve the results of TripPy that lags behind our model, we reasonably speculate that these skills can also improve the effect of LUNA.
Besides, all these models are based upon TripPy, which employs a label map as extra supervision.
The label map is a dictionary of synonyms, which is used during the testing phase.
For example, the official label of the slot \textit{``hotel-area”} in ontology is \textit{``centre”}.
But label map regards all its synonyms, such as \textit{``center”}, are also the ground truth.
We think that this manner severely reduces the difficulty of the DST task. As shown in Table \ref{main result}, the performance of TripPy degrades dramatically when the label map is removed. 
By contrast, our model does not rely on any extra information and is more generalized.


Additionally, Table \ref{main result} lists the number of trainable parameters of some baselines and our model, which illustrates that our alignment module containing multiple self-attention does not introduce large model parameters.
Compared with the baselines, the size of our model is comparable.

\begin{figure}
	\centering
	\includegraphics[width=7.5cm,height=4.2cm]{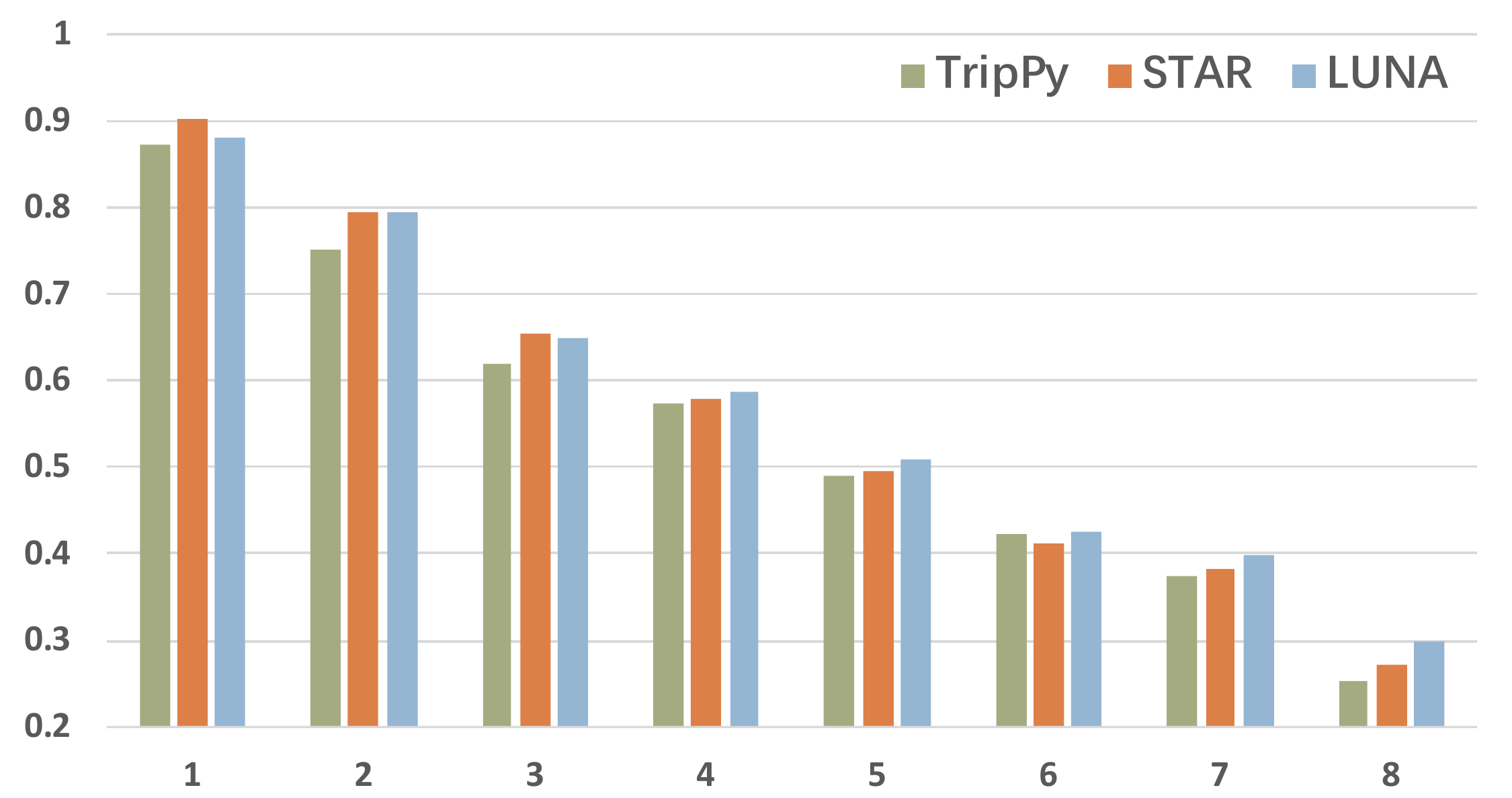}
	\caption{\label{turn} Joint accuracy at every turn.}
	\vspace{-4mm}
\end{figure}

\noindent \textbf{Accuracy at Every Turn}. In practice, the dialogue states of longer dialogues tend to be more difficult to be correctly predicted as the model needs to consider more dialogue history.
In this section, we further analyze the relationship between the depth of conversation and the prediction accuracy. 
The joint accuracy at every turn of TripPy, STAR, and LUNA on MultiWOZ 2.1 test set is shown in Figure \ref{turn}.
It presents that the scores of LUNA and STAR are basically the same when the number of conversation turns is less than 3. 
While as the conversation turns increases from 3, the superiority of LUNA gradually becomes obvious.
This is because that both TripPy and STAR exploit all dialogue utterances to assign value for each slot.
This may introduce more useless information that causes confusion to the current slots.
Whereas, LUNA only uses the most relevant utterance to assign slot value, which avoids interference by useless information.

\begin{figure*}
	\centering
	\includegraphics[width=16cm]{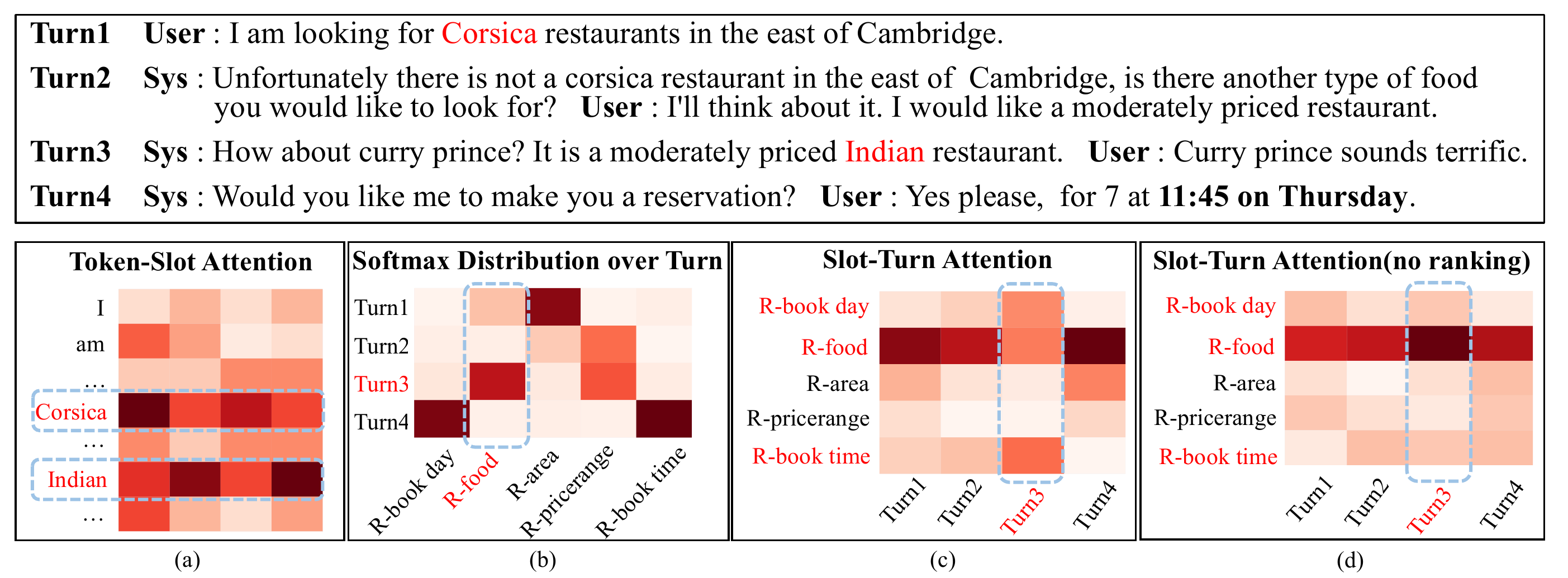}
	\caption{\label{hotmap} 
	Visualization of LUNA on an example from MultiWOZ 2.1, which is a process of  predicting the value to slot ``restaurant-food” at turn-4. ``R-" in figure denotes ``restaurant-".
	The golden value of ``restaurant-food” is ``Indian" and the confusion value is  ``Corsica". (a) is the distribution of Token-Slot Attention calculated by Eq.\ref{token-slot attention}  where the columns are the four heads in multi-head attention. (b) is the softmax distribution of alignment over turns calculated by Eq.\ref{softmax}. (c)(d) are the distributions of Slot-Turn Attention calculated by Eq.\ref{turn-slot} where (d) is the version after removing ranking loss.}
	\vspace{-3mm}
\end{figure*}

\begin{figure}[tb]
	\centering
	\includegraphics[width=8cm]{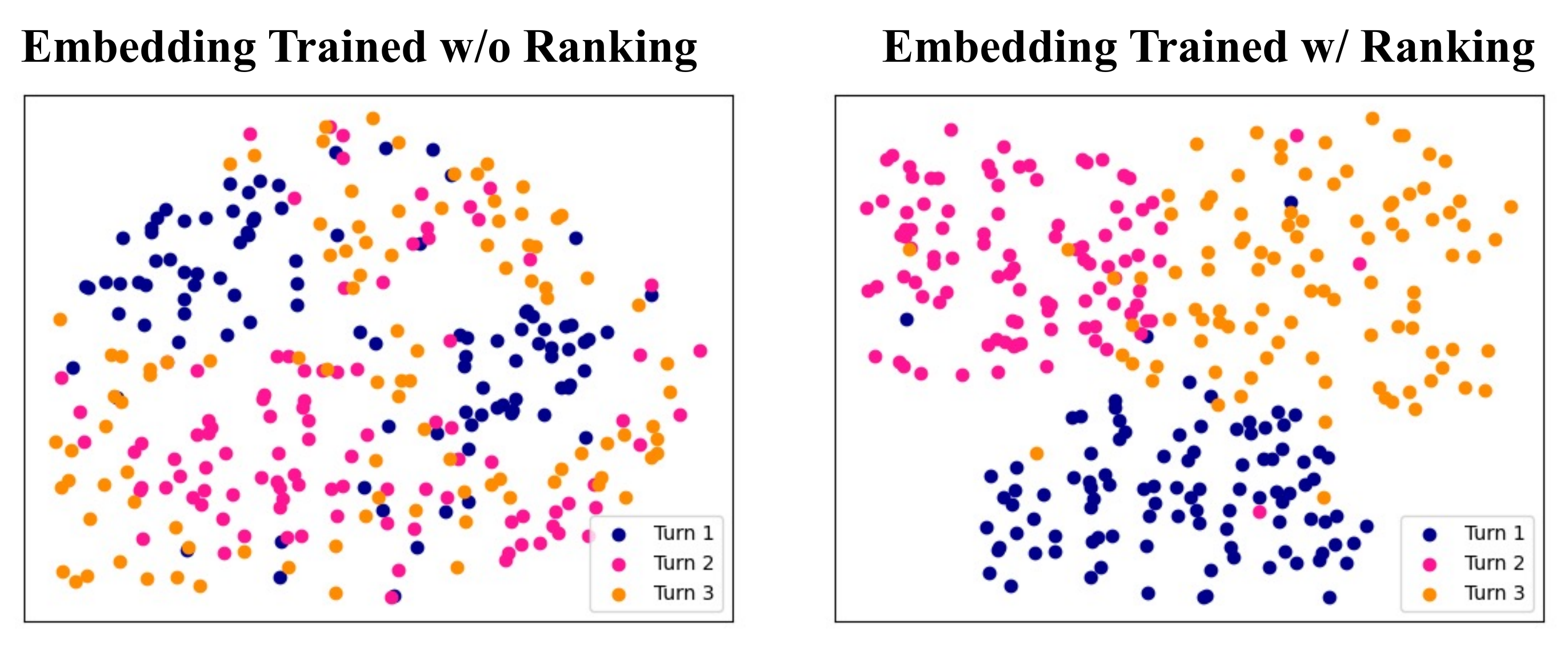}
	\caption{\label{embedding} Visualization of slot embedding using t-sne. Each point represents the slot aligned to the corresponding turn. We plot 100 slot embeddings for each turn.}
	\label{t-sne}
	\vspace{-3mm}
\end{figure}




\subsection{Ablation on Alignment Module}
To explore the effectiveness of each part in our proposed alignment module, we conduct an ablation study of these parts on the test set of MultiWOZ 2.1, as shown in Table \ref{Alignment}.

\begin{table}[h]
	\centering
	\small
	\resizebox{0.47\textwidth}{!}{
	\setlength{\tabcolsep}{1mm}
	\begin{tabular}{lll}
	    \toprule
		Model & Align Acc  & Joint Acc. \\
 		\midrule
 		LUNA   & 97.50 & 57.62  \\
 		~ - Alignment module & ~~~~-- & 53.46 (-4.16) \\
		~ - Overall slot-to-turn alignment  & 95.23 (-2.27) &  54.70 (-2.92) \\
		~ - Auxiliary task  & 96.30 (-1.20)  & 55.29 (-2.33) \\

     	\bottomrule 
	\end{tabular}
~~~	}
	\caption{\label{Alignment} The ablation study of the alignment module on the MultiWOZ 2.1. Alignment accuracy (\%) is defined as the ratio of dialogue for which the utterance turn of each slot is correctly aligned.}
\end{table}



First, we remove the whole alignment module and only use the representations of slots obtained by token-slot attention Eq.\ref{token-slot attention} to match the value.
The results show that model performance has dropped a lot (4.16 joint accuracy), proving that there are many useless tokens in the conversation history, which interfere the prediction accuracy of slot value.
Next, we remove the layer of overall slot-to-turn alignment.
We can see that this also severely damages the model performance on both alignment accuracy and joint accuracy. 
This illustrates that it is not enough to only use the information of a single slot for the alignment. 
The model needs to comprehensively consider all slots information, such as semantic correlations and temporal correlations among slots to accurately align slots and dialogue turns.
Finally, we remove the auxiliary ranking task and the results decrease by 1.20 on alignment accuracy and 2.33 on joint accuracy.
This proves that the temporal correlation among slots is important in our model which could facilitate the alignment, as we explained in the section of Introduction.
More intuitive explanations will be given in the next section through an example of visualization.

\noindent \textbf{Hard or Soft Alignment}. From Table ~\ref{Alignment}, although our hard alignment is highly accurate (Acc 97.50), we should further explore whether it can be replaced by a soft alignment to avoid the risk of error propagation.
Whereupon, we design a soft alignment that is a weighted sum of all turns with the alignment distribution over the turns as weights (Eq.~\ref{softmax}). The experimental results show that compared with hard alignment, the Joint accuracy of soft alignment on MultiWOZ 2.1 drops to 57.53 (-0.09). 
The reason is that the soft alignment encounters the problem of noises introduced from irrelevant utterances.
In other words, risk of error propagation and noise-avoiding are a trade-off.
The experimental results show that the benefits of our proposed hard alignment outweigh the risk.



\subsection{Visualization}
Figure \ref{hotmap} gives an example to visualize the process of predicting the value to slot ``restaurant-food”.
In this example,  the golden slot value is ``Indian" and ``Corsica" is the confusion value.
As shown in sub-figure (a), the slot assigns high attention weight in all heads to both ``Indian" and ``Corsica" , because as of this step, it cannot determine which one is the correct value.
At the last step, after the bi-directional fusion in our proposed alignment module, the model successfully assigns turn-3 a larger alignment score than turn-1, as shown in sub-figure (b).
In other words, the model has realized that the utterance of turn-3 (containing ``Indian") is more important than turn-1 (containing ``Corsica"). This can avoid the confusion caused by ``Corsica".

We next analyze sub-figures (c) and (d).
As we can see, all turns focus on the slot ``restaurant-food” as they incorporates its single slot information through Eq.\ref{single}.
For the column of turn-3, if the model is supervised with the auxiliary ranking task, it will also consider the information of ``restaurant-book day” and ``restaurant-book time”. 
Sub-figure (b) indicates that these two slots are easy-aligned (with turn-4).
Meanwhile, the model learns that the order of the three slots is  [``restaurant-food”, ``restaurant-book day”, ``restaurant-book time”].
Thereby, the alignment of ``restaurant-food” and turn-3 becomes easier.

Figure~\ref{t-sne} displays the 2-d visualization of slot embeddings obtained by Eq.~\ref{slot}. 
It can be seen that without ranking loss, the slot representations are irregular and borderless. Under the supervision of the ranking loss, the model can learn the boundaries between the slots aligned with different turns.

\section{Conclusion}
In this work, we reveal the problem in DST that exploiting all dialogue utterances to assign value to slots may cause suboptimal results.
To alleviate it, we propose LUNA, a slot-turn alignment enhanced approach.
and design a ranking-based auxiliary task to supervise LUNA to learn 
the temporal correlations among slots.
Comprehensive experiments are conducted on MultiWOZ 2.0, 2.1, and 2.2 and the results show that LUNA achieves new state-of-the-art results.
Moreover, the visualization demonstrates the interpretability of LUNA.

\section{Acknowledge}
We would like to thank the anonymous reviewers for their useful feedback. 
This work is supported by the National Key Research and Development Program of China under Grant No. 2020AAA0108600.

\bibliography{LUNA.new}

\begin{thebibliography}{31}
\expandafter\ifx\csname natexlab\endcsname\relax\def\natexlab#1{#1}\fi

\bibitem[{Budzianowski et~al.(2018)Budzianowski, Wen, Tseng, Casanueva, Ultes,
  Ramadan, and Ga{\v{s}}i{\'c}}]{budzianowski2018multiwoz}
Pawe{\l} Budzianowski, Tsung-Hsien Wen, Bo-Hsiang Tseng, I{\~n}igo Casanueva,
  Stefan Ultes, Osman Ramadan, and Milica Ga{\v{s}}i{\'c}. 2018.
\newblock \href {https://doi.org/10.18653/v1/D18-1547} {{M}ulti{WOZ} - a
  large-scale multi-domain {W}izard-of-{O}z dataset for task-oriented dialogue
  modelling}.
\newblock In \emph{Proceedings of the 2018 Conference on Empirical Methods in
  Natural Language Processing}, pages 5016--5026, Brussels, Belgium.
  Association for Computational Linguistics.

\bibitem[{Chen et~al.(2020{\natexlab{a}})Chen, Zhang, Mao, and
  Xu}]{chen2020parallel}
Junfan Chen, Richong Zhang, Yongyi Mao, and Jie Xu. 2020{\natexlab{a}}.
\newblock \href {https://doi.org/10.18653/v1/2020.emnlp-main.151} {Parallel
  interactive networks for multi-domain dialogue state generation}.
\newblock In \emph{Proceedings of the 2020 Conference on Empirical Methods in
  Natural Language Processing (EMNLP)}, pages 1921--1931, Online. Association
  for Computational Linguistics.

\bibitem[{Chen et~al.(2020{\natexlab{b}})Chen, Lv, Wang, Zhu, Tan, and
  Yu}]{chen2020schema}
Lu~Chen, Boer Lv, Chi Wang, Su~Zhu, Bowen Tan, and Kai Yu. 2020{\natexlab{b}}.
\newblock \href {https://aaai.org/ojs/index.php/AAAI/article/view/6250}
  {Schema-guided multi-domain dialogue state tracking with graph attention
  neural networks}.
\newblock In \emph{The Thirty-Fourth {AAAI} Conference on Artificial
  Intelligence, {AAAI} 2020, The Thirty-Second Innovative Applications of
  Artificial Intelligence Conference, {IAAI} 2020, The Tenth {AAAI} Symposium
  on Educational Advances in Artificial Intelligence, {EAAI} 2020, New York,
  NY, USA, February 7-12, 2020}, pages 7521--7528. {AAAI} Press.

\bibitem[{Dai et~al.(2021)Dai, Li, Li, Sun, Huang, Si, and
  Zhu}]{dai2021preview}
Yinpei Dai, Hangyu Li, Yongbin Li, Jian Sun, Fei Huang, Luo Si, and Xiaodan
  Zhu. 2021.
\newblock \href {https://arxiv.org/abs/2106.00291} {Preview, attend and review:
  Schema-aware curriculum learning for multi-domain dialog state tracking}.
\newblock \emph{ArXiv preprint}, abs/2106.00291.

\bibitem[{Devlin et~al.(2019)Devlin, Chang, Lee, and
  Toutanova}]{devlin2018bert}
Jacob Devlin, Ming-Wei Chang, Kenton Lee, and Kristina Toutanova. 2019.
\newblock \href {https://doi.org/10.18653/v1/N19-1423} {{BERT}: Pre-training of
  deep bidirectional transformers for language understanding}.
\newblock In \emph{Proceedings of the 2019 Conference of the North {A}merican
  Chapter of the Association for Computational Linguistics: Human Language
  Technologies, Volume 1 (Long and Short Papers)}, pages 4171--4186,
  Minneapolis, Minnesota. Association for Computational Linguistics.

\bibitem[{Eric et~al.(2019)Eric, Goel, Paul, Sethi, Agarwal, Gao, and
  Hakkani-T{\"u}r}]{eric2019multiwoz}
Mihail Eric, Rahul Goel, Shachi Paul, Abhishek Sethi, Sanchit Agarwal, Shuyang
  Gao, and Dilek Hakkani-T{\"u}r. 2019.
\newblock Multiwoz 2.1: Multi-domain dialogue state corrections and state
  tracking baselines.

\bibitem[{Feng et~al.(2021)Feng, Wang, and Li}]{feng2020sequence}
Yue Feng, Yang Wang, and Hang Li. 2021.
\newblock \href {https://doi.org/10.18653/v1/2021.acl-long.135} {A
  sequence-to-sequence approach to dialogue state tracking}.
\newblock In \emph{Proceedings of the 59th Annual Meeting of the Association
  for Computational Linguistics and the 11th International Joint Conference on
  Natural Language Processing (Volume 1: Long Papers)}, pages 1714--1725,
  Online. Association for Computational Linguistics.

\bibitem[{Heck et~al.(2020)Heck, van Niekerk, Lubis, Geishauser, Lin, Moresi,
  and Gasic}]{heck2020trippy}
Michael Heck, Carel van Niekerk, Nurul Lubis, Christian Geishauser, Hsien-Chin
  Lin, Marco Moresi, and Milica Gasic. 2020.
\newblock \href {https://aclanthology.org/2020.sigdial-1.4} {{T}rip{P}y: A
  triple copy strategy for value independent neural dialog state tracking}.
\newblock In \emph{Proceedings of the 21th Annual Meeting of the Special
  Interest Group on Discourse and Dialogue}, pages 35--44, 1st virtual meeting.
  Association for Computational Linguistics.

\bibitem[{Hosseini{-}Asl et~al.(2020)Hosseini{-}Asl, McCann, Wu, Yavuz, and
  Socher}]{hosseini2020simple}
Ehsan Hosseini{-}Asl, Bryan McCann, Chien{-}Sheng Wu, Semih Yavuz, and Richard
  Socher. 2020.
\newblock \href
  {https://proceedings.neurips.cc/paper/2020/hash/e946209592563be0f01c844ab2170f0c-Abstract.html}
  {A simple language model for task-oriented dialogue}.
\newblock In \emph{Advances in Neural Information Processing Systems 33: Annual
  Conference on Neural Information Processing Systems 2020, NeurIPS 2020,
  December 6-12, 2020, virtual}.

\bibitem[{Hu et~al.(2020)Hu, Yang, Chen, He, and Yu}]{hu2020sas}
Jiaying Hu, Yan Yang, Chencai Chen, Liang He, and Zhou Yu. 2020.
\newblock \href {https://doi.org/10.18653/v1/2020.acl-main.567} {{SAS}:
  Dialogue state tracking via slot attention and slot information sharing}.
\newblock In \emph{Proceedings of the 58th Annual Meeting of the Association
  for Computational Linguistics}, pages 6366--6375, Online. Association for
  Computational Linguistics.

\bibitem[{Kim et~al.(2020)Kim, Yang, Kim, and Lee}]{kim2019efficient}
Sungdong Kim, Sohee Yang, Gyuwan Kim, and Sang-Woo Lee. 2020.
\newblock \href {https://doi.org/10.18653/v1/2020.acl-main.53} {Efficient
  dialogue state tracking by selectively overwriting memory}.
\newblock In \emph{Proceedings of the 58th Annual Meeting of the Association
  for Computational Linguistics}, pages 567--582, Online. Association for
  Computational Linguistics.

\bibitem[{Kingma and Ba(2015)}]{kingma2014adam}
Diederik~P. Kingma and Jimmy Ba. 2015.
\newblock \href {http://arxiv.org/abs/1412.6980} {Adam: {A} method for
  stochastic optimization}.
\newblock In \emph{3rd International Conference on Learning Representations,
  {ICLR} 2015, San Diego, CA, USA, May 7-9, 2015, Conference Track
  Proceedings}.

\bibitem[{Lee et~al.(2019)Lee, Lee, and Kim}]{lee2019sumbt}
Hwaran Lee, Jinsik Lee, and Tae-Yoon Kim. 2019.
\newblock \href {https://doi.org/10.18653/v1/P19-1546} {{SUMBT}: Slot-utterance
  matching for universal and scalable belief tracking}.
\newblock In \emph{Proceedings of the 57th Annual Meeting of the Association
  for Computational Linguistics}, pages 5478--5483, Florence, Italy.
  Association for Computational Linguistics.

\bibitem[{Li et~al.(2020)Li, Yavuz, Hashimoto, Li, Niu, Rajani, Yan, Zhou, and
  Xiong}]{li2020coco}
Shiyang Li, Semih Yavuz, Kazuma Hashimoto, Jia Li, Tong Niu, Nazneen Rajani,
  Xifeng Yan, Yingbo Zhou, and Caiming Xiong. 2020.
\newblock \href {https://arxiv.org/abs/2010.12850} {Coco: Controllable
  counterfactuals for evaluating dialogue state trackers}.
\newblock \emph{ArXiv preprint}, abs/2010.12850.

\bibitem[{Mehri et~al.(2020)Mehri, Eric, and Hakkani-Tur}]{mehri2020dialoglue}
Shikib Mehri, Mihail Eric, and Dilek Hakkani-Tur. 2020.
\newblock \href {https://arxiv.org/abs/2009.13570} {Dialoglue: A natural
  language understanding benchmark for task-oriented dialogue}.
\newblock \emph{ArXiv preprint}, abs/2009.13570.

\bibitem[{Mrk{\v{s}}i{\'c} et~al.(2017)Mrk{\v{s}}i{\'c}, {\'O}~S{\'e}aghdha,
  Wen, Thomson, and Young}]{mrkvsic2016neural}
Nikola Mrk{\v{s}}i{\'c}, Diarmuid {\'O}~S{\'e}aghdha, Tsung-Hsien Wen, Blaise
  Thomson, and Steve Young. 2017.
\newblock \href {https://doi.org/10.18653/v1/P17-1163} {Neural belief tracker:
  Data-driven dialogue state tracking}.
\newblock In \emph{Proceedings of the 55th Annual Meeting of the Association
  for Computational Linguistics (Volume 1: Long Papers)}, pages 1777--1788,
  Vancouver, Canada. Association for Computational Linguistics.

\bibitem[{Quan and Xiong(2020)}]{quan2020modeling}
Jun Quan and Deyi Xiong. 2020.
\newblock \href {https://doi.org/10.18653/v1/2020.acl-main.637} {Modeling long
  context for task-oriented dialogue state generation}.
\newblock In \emph{Proceedings of the 58th Annual Meeting of the Association
  for Computational Linguistics}, pages 7119--7124, Online. Association for
  Computational Linguistics.

\bibitem[{Ren et~al.(2018)Ren, Xie, Chen, and Yu}]{ren2018towards}
Liliang Ren, Kaige Xie, Lu~Chen, and Kai Yu. 2018.
\newblock \href {https://doi.org/10.18653/v1/D18-1299} {Towards universal
  dialogue state tracking}.
\newblock In \emph{Proceedings of the 2018 Conference on Empirical Methods in
  Natural Language Processing}, pages 2780--2786, Brussels, Belgium.
  Association for Computational Linguistics.

\bibitem[{Shan et~al.(2020)Shan, Li, Zhang, Meng, Feng, Niu, and
  Zhou}]{shan2020contextual}
Yong Shan, Zekang Li, Jinchao Zhang, Fandong Meng, Yang Feng, Cheng Niu, and
  Jie Zhou. 2020.
\newblock \href {https://doi.org/10.18653/v1/2020.acl-main.563} {A contextual
  hierarchical attention network with adaptive objective for dialogue state
  tracking}.
\newblock In \emph{Proceedings of the 58th Annual Meeting of the Association
  for Computational Linguistics}, pages 6322--6333, Online. Association for
  Computational Linguistics.

\bibitem[{Sun et~al.(2022)Sun, Bao, Wu, and He}]{BORT}
Haipeng Sun, Junwei Bao, Youzheng Wu, and Xiaodong He. 2022.
\newblock Bort: Back and denoising reconstruction for end-to-end task-oriented
  dialog.
\newblock In \emph{Findings of the Association for Computational Linguistics:
  NAACL 2022}. Association for Computational Linguistics.

\bibitem[{Vaswani et~al.(2017)Vaswani, Shazeer, Parmar, Uszkoreit, Jones,
  Gomez, Kaiser, and Polosukhin}]{vaswani2017attention}
Ashish Vaswani, Noam Shazeer, Niki Parmar, Jakob Uszkoreit, Llion Jones,
  Aidan~N. Gomez, Lukasz Kaiser, and Illia Polosukhin. 2017.
\newblock \href
  {https://proceedings.neurips.cc/paper/2017/hash/3f5ee243547dee91fbd053c1c4a845aa-Abstract.html}
  {Attention is all you need}.
\newblock In \emph{Advances in Neural Information Processing Systems 30: Annual
  Conference on Neural Information Processing Systems 2017, December 4-9, 2017,
  Long Beach, CA, {USA}}, pages 5998--6008.

\bibitem[{Williams et~al.(2016)Williams, Raux, and
  Henderson}]{williams2016dialog}
Jason~D Williams, Antoine Raux, and Matthew Henderson. 2016.
\newblock The dialog state tracking challenge series: A review.
\newblock \emph{Dialogue \& Discourse}, 7(3):4--33.

\bibitem[{Wu et~al.(2019)Wu, Madotto, Hosseini-Asl, Xiong, Socher, and
  Fung}]{wu2019transferable}
Chien-Sheng Wu, Andrea Madotto, Ehsan Hosseini-Asl, Caiming Xiong, Richard
  Socher, and Pascale Fung. 2019.
\newblock \href {https://doi.org/10.18653/v1/P19-1078} {Transferable
  multi-domain state generator for task-oriented dialogue systems}.
\newblock In \emph{Proceedings of the 57th Annual Meeting of the Association
  for Computational Linguistics}, pages 808--819, Florence, Italy. Association
  for Computational Linguistics.

\bibitem[{Xia et~al.(2008)Xia, Liu, Wang, Zhang, and Li}]{xia2008listwise}
Fen Xia, Tie{-}Yan Liu, Jue Wang, Wensheng Zhang, and Hang Li. 2008.
\newblock \href {https://doi.org/10.1145/1390156.1390306} {Listwise approach to
  learning to rank: theory and algorithm}.
\newblock In \emph{Machine Learning, Proceedings of the Twenty-Fifth
  International Conference {(ICML} 2008), Helsinki, Finland, June 5-9, 2008},
  volume 307 of \emph{{ACM} International Conference Proceeding Series}, pages
  1192--1199. {ACM}.

\bibitem[{Xie et~al.(2018)Xie, Chang, Ren, Chen, and Yu}]{xie2018cost}
Kaige Xie, Cheng Chang, Liliang Ren, Lu~Chen, and Kai Yu. 2018.
\newblock \href {https://doi.org/10.18653/v1/W18-5022} {Cost-sensitive active
  learning for dialogue state tracking}.
\newblock In \emph{Proceedings of the 19th Annual {SIG}dial Meeting on
  Discourse and Dialogue}, pages 209--213, Melbourne, Australia. Association
  for Computational Linguistics.

\bibitem[{Xu and Hu(2018)}]{xu2018end}
Puyang Xu and Qi~Hu. 2018.
\newblock \href {https://doi.org/10.18653/v1/P18-1134} {An end-to-end approach
  for handling unknown slot values in dialogue state tracking}.
\newblock In \emph{Proceedings of the 56th Annual Meeting of the Association
  for Computational Linguistics (Volume 1: Long Papers)}, pages 1448--1457,
  Melbourne, Australia. Association for Computational Linguistics.

\bibitem[{Ye et~al.(2021)Ye, Manotumruksa, Zhang, Li, and Yilmaz}]{ye2021slot}
Fanghua Ye, Jarana Manotumruksa, Qiang Zhang, Shenghui Li, and Emine Yilmaz.
  2021.
\newblock Slot self-attentive dialogue state tracking.
\newblock In \emph{Proceedings of the Web Conference 2021}, pages 1598--1608.

\bibitem[{Zang et~al.(2020)Zang, Rastogi, Sunkara, Gupta, Zhang, and
  Chen}]{zang2020multiwoz}
Xiaoxue Zang, Abhinav Rastogi, Srinivas Sunkara, Raghav Gupta, Jianguo Zhang,
  and Jindong Chen. 2020.
\newblock \href {https://doi.org/10.18653/v1/2020.nlp4convai-1.13}
  {{M}ulti{WOZ} 2.2 : A dialogue dataset with additional annotation corrections
  and state tracking baselines}.
\newblock In \emph{Proceedings of the 2nd Workshop on Natural Language
  Processing for Conversational AI}, pages 109--117, Online. Association for
  Computational Linguistics.

\bibitem[{Zhang et~al.(2020)Zhang, Hashimoto, Wu, Wang, Yu, Socher, and
  Xiong}]{zhang2019find}
Jianguo Zhang, Kazuma Hashimoto, Chien-Sheng Wu, Yao Wang, Philip Yu, Richard
  Socher, and Caiming Xiong. 2020.
\newblock \href {https://aclanthology.org/2020.starsem-1.17} {Find or classify?
  dual strategy for slot-value predictions on multi-domain dialog state
  tracking}.
\newblock In \emph{Proceedings of the Ninth Joint Conference on Lexical and
  Computational Semantics}, pages 154--167, Barcelona, Spain (Online).
  Association for Computational Linguistics.

\bibitem[{Zhong et~al.(2018)Zhong, Xiong, and Socher}]{zhong2018global}
Victor Zhong, Caiming Xiong, and Richard Socher. 2018.
\newblock \href {https://doi.org/10.18653/v1/P18-1135} {Global-locally
  self-attentive encoder for dialogue state tracking}.
\newblock In \emph{Proceedings of the 56th Annual Meeting of the Association
  for Computational Linguistics (Volume 1: Long Papers)}, pages 1458--1467,
  Melbourne, Australia. Association for Computational Linguistics.

\bibitem[{Zhu et~al.(2020)Zhu, Li, Chen, and Yu}]{zhu2020efficient}
Su~Zhu, Jieyu Li, Lu~Chen, and Kai Yu. 2020.
\newblock \href {https://doi.org/10.18653/v1/2020.findings-emnlp.68} {Efficient
  context and schema fusion networks for multi-domain dialogue state tracking}.
\newblock In \emph{Findings of the Association for Computational Linguistics:
  EMNLP 2020}, pages 766--781, Online. Association for Computational
  Linguistics.

\end{thebibliography}
\bibliographystyle{acl_natbib}




\end{document}